\documentclass[10pt,twocolumn,letterpaper]{article}

\usepackage{cvpr}
\usepackage{times}
\usepackage{epsfig}
\usepackage{graphicx}
\usepackage{amsmath}
\usepackage{amssymb}
\usepackage{algorithm}
\usepackage{algorithmic}
\usepackage{multirow}
\usepackage{makecell}
\usepackage{booktabs}

\usepackage[breaklinks=true,bookmarks=false]{hyperref}

\cvprfinalcopy 

\def\cvprPaperID{4251} 

\begin{document}

\title{AdversarialNAS: Adversarial Neural Architecture Search for GANs}

\author{
  Chen Gao$^{1,2,4}$,\quad Yunpeng Chen$^{4}$,\quad Si Liu$^{3}$\thanks{Corresponding author},\quad Zhenxiong Tan$^{5}$,\quad Shuicheng Yan$^{4}$ \\
  $^{1}$Institute of Information Engineering, Chinese Academy of Sciences \\
  $^{2}$University of Chinese Academy of Sciences \\
  $^{3}$School of Computer Science and Engineering, Beihang University \\
  $^{4}$Yitu Technology \quad
  $^{5}$Beijing Forestry University \\
  {\tt\small gaochen@iie.ac.cn, liusi@buaa.edu.cn, yunpeng.chen@yitu-inc.com}
}


\maketitle
\thispagestyle{empty}

\begin{abstract}
Neural Architecture Search (NAS) that aims to automate the procedure of architecture design has achieved promising results in many computer vision fields. In this paper, we propose an AdversarialNAS method specially tailored for Generative Adversarial Networks (GANs) to search for a superior generative model on the task of unconditional image generation. The AdversarialNAS is the first method that can search the architectures of generator and discriminator simultaneously in a differentiable manner. During searching, the designed adversarial search algorithm does not need to comput any extra metric to evaluate the performance of the searched architecture, and the search paradigm considers the relevance between the two network architectures and improves their mutual balance. Therefore, AdversarialNAS is very efficient and only takes 1 GPU day to search for a superior generative model in the proposed large search space ($10^{38}$). Experiments demonstrate the effectiveness and superiority of our method. The discovered generative model sets a new state-of-the-art FID score of $10.87$ and highly competitive Inception Score of $8.74$ on CIFAR-10. Its transferability is also proven by setting new state-of-the-art FID score of $26.98$ and Inception score of $9.63$ on STL-10. Code is at: \url{https://github.com/chengaopro/AdversarialNAS}.
\end{abstract}

\vspace{-3mm}
\section{Introduction}
\begin{figure}
   \begin{center} 
   \includegraphics[width=1\linewidth]{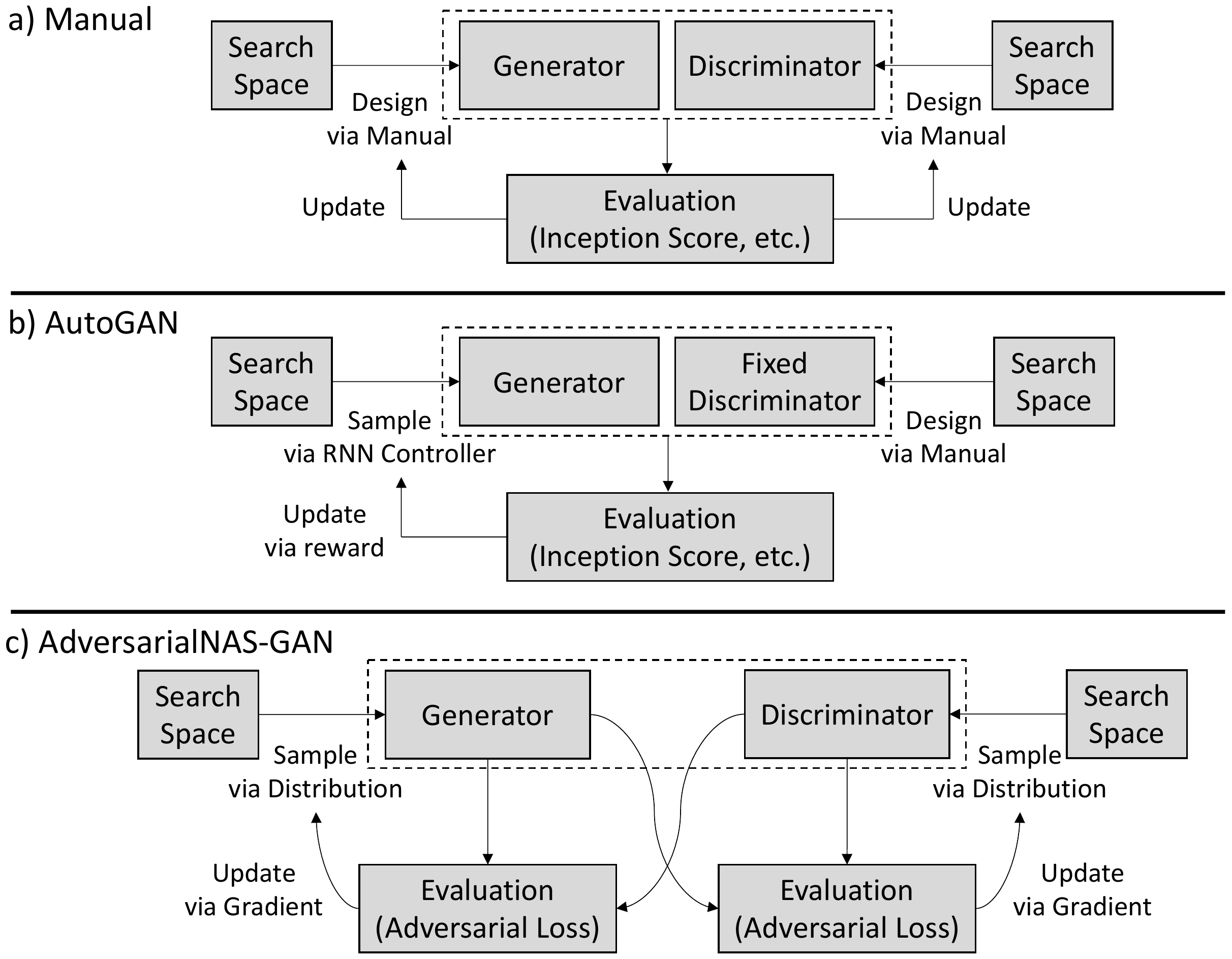}
\end{center}
   \caption{Comparisons of different ways of designing GAN architectures. a) The previous hand-crafted GAN architectures depend on the experience of human experts. b) AutoGAN~\cite{gong2019autogan} adopts IS or FID as reward to update the architecture controller via reinforcement learning. c) The proposed AdversarialNAS searches architecture in a differentiable way with an adversarial search mechanism, which achieves better performance with higher efficiency.}
   \label{fig:compare}
   \vspace{-4mm}
\end{figure}

Image generation is a fundamental task in the field of computer vision. Recently, GANs~\cite{goodfellow2014generative} have attracted much attention due to their remarkable performance for generating realistic images.
Previous architectures of GANs are designed by human experts with laborious trial-and-error testings (Fig.~\ref{fig:compare} a)) and the instability issue in GAN training extremely increases the difficulty of architecture design. Therefore, the architecture of the generative model in GAN literature has very few types and can be simply divided into two styles: DCGANs-based~\cite{radford2015unsupervised} and ResNet-based~\cite{he2016deep}.
On the other hand, the benefits of specially designing the network architecture have been proven through lost of discriminative networks, such as ResNet~\cite{he2016deep}, DenseNet~\cite{huang2017densely}, MobileNet~\cite{sandler2018mobilenetv2}, ShuffleNet~\cite{zhang2018shufflenet}, EfficientNet \cite{tan2019efficientnet}, HRNet~\cite{sun2019deep} and \cite{chen2019drop,gao2020highly}. Therefore, the research about the backbone architecture of GANs needs more attention to further improve the performance of the generative model.


Recently, Neural Architecture Search (NAS) has been studied heatedly owing to its ability to automatically discover the optimal network architecture, which significantly reduces human labor. However, on generation tasks, specifically GANs-based generation, only AutoGAN~\cite{gong2019autogan} and AGAN~\cite{wang2019agan} have explored the application of NAS. 

To design a NAS algorithm specially tailored for GANs on the unconditional image generation task, there are two main challenges. First, it is expected to utilize an efficient supervision signal to guide the search process in this unsupervised task. However, the existing works~\cite{gong2019autogan,wang2019agan} both adopt the Inception Score (IS)~\cite{salimans2016improved} or FID to evaluate the architecture performance and take IS or FID as a reward to update the architecture controller via reinforcement learning strategy. Obtaining IS or FID needs to generate hundreds of images and use the statistics produced by an Inception network to calculate the final score. Thus it is extremely time-consuming, e.g. 200 GPUs over 6 days~\cite{wang2019agan}. Second, the relevance and balance between generator and discriminator need to be considered during searching since the training process of GANs is a unique competition. However, AutoGAN search for a generator with a pre-defined growing discriminator (Fig.~\ref{fig:compare} b)), where the architecture of the discriminator can be regarded as fixed and may limit the algorithm to search for an optimal architecture of generator.


In this work, we propose an \textbf{Adversarial} \textbf{N}eural \textbf{A}rchitecture \textbf{S}earch (AdversarialNAS) method to address the above challenges (Fig.~\ref{fig:compare} c)). First, we design a large search space ($10^{38}$) for fragile GAN and relax the search space to be continuous. Thus the architecture can be represented by a set of continuous variables obeying certain probability distribution and searched in a differentiable manner. Second, we propose to directly utilize the existing discriminator to evaluate the architecture of generator in each search iteration. Specifically, when searching for the generator architecture, the discriminator provides the supervision signal to guide the search direction, which is technically utilized to update the architecture distribution of generator through gradient descent. Therefore, our method is much more efficient since the extra computation cost for calculating evaluation metric is eliminated. Third, in order to consider the relevance and balance between the generator and discriminator, we propose to dynamically change the architecture of discriminator simultaneously during searching. Accordingly, we adopt the generator to evaluate the architecture of discriminator and comput the loss to update the discriminator architecture through ascending the stochastic gradient. The two architectures play against each other in a competition to continually improve their performance, which is essentially an adversarial searching mechanism. Therefore, the AdversarialNAS gets rid of calculating extra evaluation metric and solves the unsupervised task through an adversarial mechanism. It adequately considers the mutual balance between the two architectures, which benefits for searching a superior generative model.


To sum up, our main contributions are three-fold.
\begin{itemize}
   \vspace{-1mm}
   \item We propose a novel AdversarialNAS method, which is the first gradient-based NAS method in GAN field and achieves state-of-art performance with much higher efficiency. 
   We design a large architecture search space ($10^{38}$) for GAN and make it feasible to search in. 
   Our AdversarialNAS can only tasks 1 GPU day for searching an optimal architecture in the large search space.
   \item Considering GAN is an unique competition between two networks, the proposed AdversarialNAS alternately searches the architecture of both of them under an adversarial searching strategy to improve their mutual balance, which is specifically tailored for GAN.
   \item The searched architecture has more advanced transferability and scalability while achieving state-of-the-art performance on both CIFAR-10 and STL-10 datasets.
\end{itemize}

\section{Related Work}
\subsection{Generative Adversarial Networks}
Although Restricted Boltzmann Machines~\cite{hinton2006reducing} and flow-based generative models~\cite{dinh2014nice} are all capable of generating natural images, GANs~\cite{goodfellow2014generative} are still the most widely used methods in recent years due to their impressive generation ability. GANs based approaches have achieved advanced results in various generation tasks, such as image-to-image translation \cite{isola2017image,choi2018stargan,Jiang2019PSGANPS,zhu2019ugan}, text-to-image translation~\cite{xu2018attngan,zhang2017stackgan} and image inpainting~\cite{pathak2016context}. However, the potential of GANs has not been fully explored since there is rare work~\cite{radford2015unsupervised} studying the impact of architecture design on the performance of GANs. In this work, we aim to search for a powerful and effective network structure specifically for the generative model via an automatic manner.

\subsection{Neural Architecture Search}
Automatic Machine Learning (AutoML) has attracted lots of attention recently, and Neural Architecture Search (NAS) is one of the most important direction. The goal of NAS is to automatically search for an effective architecture that satisfies certain demands. The NAS technique has applied to many computer vision tasks such as image classification~\cite{cai2018proxylessnas,liu2018progressive,liu2018darts,pham2018efficient,zoph2016neural,chen2019efficient}, dense image prediction~\cite{liu2019auto,zhang2019customizable,chen2018searching} and object detection~\cite{ghiasi2019fpn,peng2019efficient}.

Early works of NAS adopt heuristic methods such as reinforcement learning~\cite{zoph2016neural} and evolutionary algorithm~\cite{xie2017genetic}. Obtaining an architecture with remarkable performance using such methods requires huge computational resources, e.g., 2000 GPUs days~\cite{xie2017genetic}. Therefore, lots of works design various strategies to reduce the expensive costs including weight sharing~\cite{pham2018efficient}, performance prediction~\cite{brock2018smash}, progressive manner~\cite{liu2018progressive} and one-shot mechanism~\cite{liu2018darts,xie2018snas}. The DARTS~\cite{liu2018darts} in one-shot literature is the first approach that relaxes the search space to be continuous and conducts searching in a differentiable way. The architecture parameters and network weights can be trained simultaneously in an end-to-end fashion by gradient descent. Thus it extremely compresses the search time.

However, all of these methods are designed for recognition and supervision tasks. 
To the best of our knowledge, there have been limited works~\cite{gong2019autogan} exploring applying NAS to unsupervised or weakly supervised tasks. In this work, we present the first gradient-based NAS method in GAN field and achieve state-of-the-art performance with much higher efficiency in the unsupervised image generation task.

\subsection{NAS in GANs}
Recently, a few works have attempted to incorporate neural architecture search with GANs. AutoGAN \cite{gong2019autogan} adopts the reinforcement learning strategy to discover the architecture of generative models automatically. However, it only searches for the generator with a fixed architecture of discriminator. This mechanism limits the performance of the searched generator since the stability of GANs training is highly dependent on the balance between these two players. Besides, the search space is relatively small ($10^5$), thus its randomly searched architecture can achieve acceptable results, e.g., FID (lower is better): 21.39 (random) and 12.42 (search) in CIFAR-10 dataset. The AGAN~\cite{wang2019agan} enlarges the search space specifically for the generative model, but the computational cost is expensive (1200 GPUs days) under the reinforcement learning framework. The performance of the discovered model is slightly worse, e.g., FID: 30.5 in CIFAR-10. Moreover, the reward used to update the weights of the network controller during evaluation stage is Inception Score, which is not a suitable supervisory single to guide the architecture search since it is time-consuming.

Instead, we search the architecture in a differentiable way and discard the evaluation stage. The reward of previous methods is obtained after a prolonged training and evaluation process, while our signal (loss) for guiding the search direction is given instantly in each iteration. Thus our method is more efficient. The designed adversarial search algorithm improves the mutual balance of the two networks for stabling and optimizing the search process.


\section{Method}
In this section, we first introduce the proposed search space of GANs and the way for relaxing it to be continuous. Then we describe the AdversarialNAS method.
\subsection{Search Space for GANs}

The goal of the proposed AdversarialNAS is to automatically search for an superior architecture of generative model through an adversarial searching mannr. Specifically, we aim to search for a series of cells, including Up-Cell and Down-Cell, as the building blocks to construct the final architecture of GAN. Three Up-Cells and four Down-Cells are stacked to form a generator and discriminator respectively. Since the convolution neural network has a natural hierarchical structure and each layer has unique function, we search for the cells each with a different architecture.

\begin{figure}
   \begin{center}
   \includegraphics[width=1\linewidth]{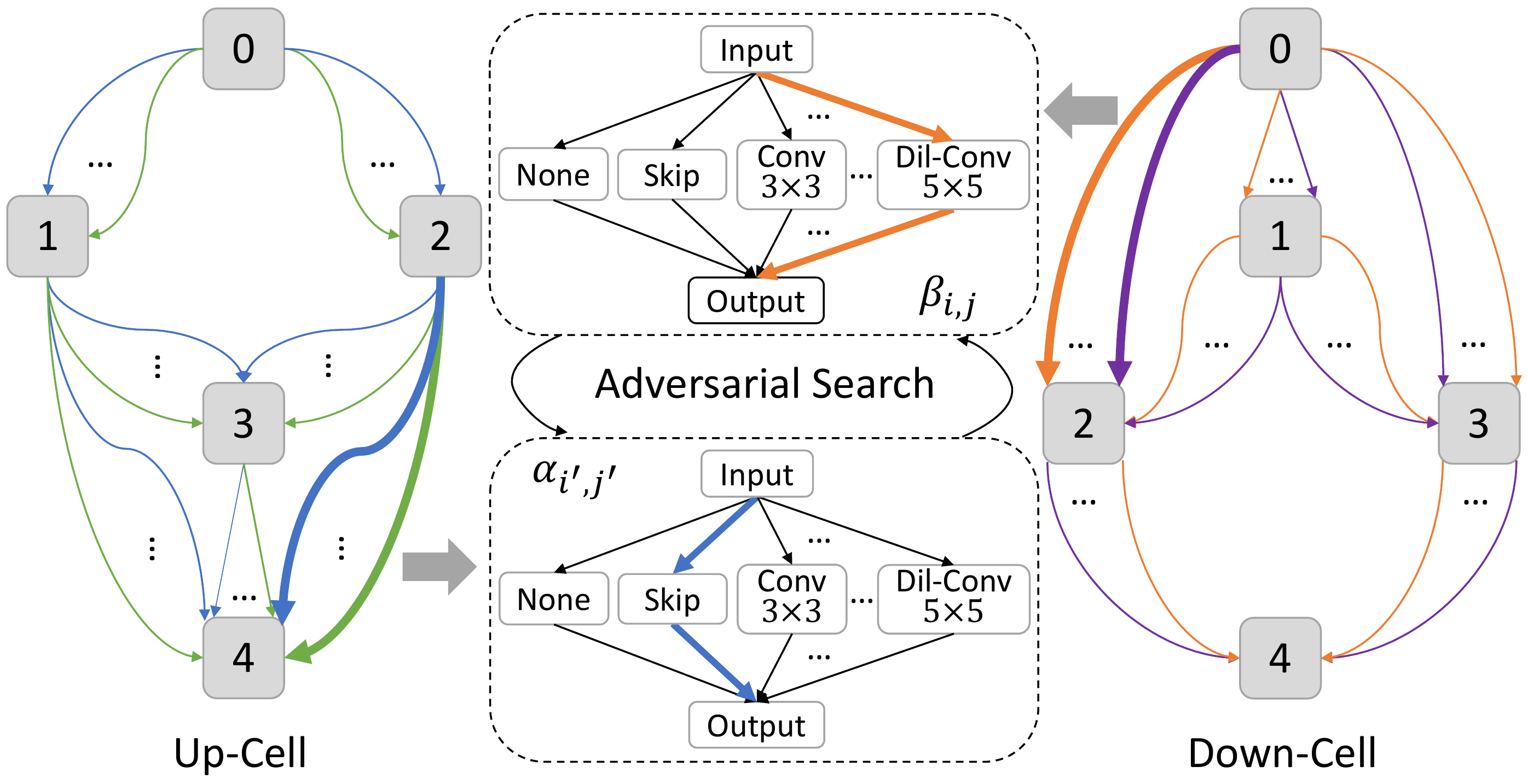}
\end{center} 
\vspace{-2.5mm}
   \caption{The search space of Up-cell and Down-Cell. The architectures of both Up-Cell and Down-Cell will continuously promote each other in an adversarial manner.}
   \label{fig:cell}
   \vspace{-3mm}
\end{figure}

We represent a cell as a Directed Acyclic Graph (DAG) consisting of an ordered sequence of N nodes (Fig.~\ref{fig:cell}). The cell takes image features as input and outputs processed features, where each node $x_i$ in DAG indicates an intermediate feature and each edge $f_{i,j}$ between two nodes $x_i, x_j$ is a specific operation. Since we aim to search for an optimal architecture of generator that is actually an upsampling network, we design a search space for specific Up-Cell that is almost fully connected topology, as given in the left of Fig.~\ref{fig:cell}. The Up-Cell consists of 4 nodes, and each node can be obtained by its previous nodes through selecting an operation from a candidate set according to the search algorithm. The search space of generator $\mathbb{F}_G$ includes a candidate set of normal operations, which is designed as below.

\begin{table}[H]
   \footnotesize
   \centering
   \vspace{-0.1in}
   \begin{tabular}{ll}
   $\bullet\;$ None                           & $\bullet\;$ Identity                           \\
   $\bullet\;$ Convolution 1x1, Dilation=1                                                      \\
   $\bullet\;$ Convolution 3x3, Dilation=1     & $\bullet\;$ Convolution 3x3, Dilation=2  \\
   $\bullet\;$ Convolution 5x5, Dilation=1     & $\bullet\;$ Convolution 5x5, Dilation=2 \\
   \end{tabular}
   \vspace{-0.15in}
\end{table} 
The `None' means there is no operation between two corresponding nodes, which is used to change the topology of the cell. The `Identity' denotes the skip connection operation that provides multi-scale features. The stride of these operations is 1 so that they will keep the spatial resolution. The search space of generator also contains a subset of upsampling operations, which is devised as below.
\begin{table}[H]
   \footnotesize
   \centering
   \vspace{-0.08in}
   \begin{tabular}{ll}
   $\bullet\;$ Transposed Convolution 3x3                                                   \\
   $\bullet\;$ Nearest Neighbor Interpolation & $\bullet\;$ Bilinear Interpolation             \\
   \end{tabular}
\end{table}
\vspace{-0.16in}
Note that, these operations can only be searched by edge $0\rightarrow 1$ and $0\rightarrow 2$ in a specific Up-Cell. To search for the generator in an adversarial way, we simply invert the Up-Cell to form a Down-Cell (shown in the right of Fig.~\ref{fig:cell}) ensuring their balance. The search space of discriminator $\mathbb{F}_D$ also contains a candidate set of normal operations, which is the same as the one of Up-Cell. However, the candidate set of downsampling operations is achieved by
\begin{table}[H]
   \footnotesize
   \centering
   \vspace{-0.1in}
   \begin{tabular}{ll}
   $\bullet\;$ Average Pooling                           & $\bullet\;$ Max Pooling                           \\
   $\bullet\;$ Convolution 3x3, Dilation=1     & $\bullet\;$ Convolution 3x3, Dilation=2  \\
   $\bullet\;$ Convolution 5x5, Dilation=1     & $\bullet\;$ Convolution 5x5, Dilation=2 \\
   \end{tabular}
   \vspace{-0.15in}
\end{table}
With stride equaling 2, the downsampling operations can only be searched in edge $2\rightarrow 4$ and $3\rightarrow 4$. Therefore, during searching, there are totally $10^{38}$ different network architectures for GANs.

\subsection{Continuous Relaxation of Architectures}
The goal of the search algorithm is to select a specific operation from the pre-defined candidate set for each edge. Therefore, the intermediate node $x_{n,j}$ in the n-th cell can be calculated through the selected functions and its previous connected nodes as $x_{n,j}=\sum_{i<j}f_{n,i,j}(x_{n,i})$. For RL-based NAS algorithms, the function $f_{n,i,j}$ is directly sampled from the candidate set according to the learnable architecture controller. Inspired by Gradient-based NAS algorithm~\cite{liu2018darts}, we relax the function $f_{n,i,j}$ to a soft version through Gumbel-Max trick~\cite{maddison2014sampling}: 
\begin{equation}
   f_{n,i,j}^{soft}(x)=\sum_{f\in \mathbb{F}_G}\frac{\exp((p_{n,i,j}^f+o^f)/ \tau)}{\sum_{f'\in \mathbb{F}_G}\exp((p_{n,i,j}^{f'}+o^{f'})/ \tau)}f(x), 
   \label{eq:architecture_g}
\end{equation}
where $o^f$ is the noise sampled from the Gumbel (0,1) distribution, and the $\tau$ is the softmax temperature. The $p_{n,i,j}^f$ is the probability of selecting a specific function $f$ in edge $i\rightarrow j$ of n-th cell. The Gumbel version softmax is applied to follow the learned probability distribution more strictly. Therefore, each edge will contain a probability vector $[p^{f_1}, ..., p^{f_m}], m=|\mathbb{F}_G|$. This discrete probability distribution is calculated through a simple softmax function as $p^f=\frac{\exp(\alpha^f)}{\sum_{f\in \mathbb{F}_G}\exp(\alpha^{f'})}$, where the $\alpha$ is the learnable parameter. Therefore, the goal of searching for an architecture is converted to learning an optimal set of probability vectors for every edge, and the architecture can be derived from the learned probability distribution. Besides, in order to dynamically change the architecture of discriminator simultaneously, we also conduct a set of continuous parameters $\beta$ for calculating the probability of each function in discriminator as $q^f=\frac{\exp(\beta^f)}{\sum_{f\in \mathbb{F}_D}\exp(\beta^{f'})}$. Therefore, the soft version of the function can be achieved like the generator as
\begin{equation}
   f_{n,i,j}^{soft}(x)=\sum_{f\in \mathbb{F}_D}\frac{\exp((q_{n,i,j}^f+o^f)/ \tau)}{\sum_{f'\in \mathbb{F}_D}\exp((q_{n,i,j}^{f'}+o^{f'})/ \tau)}f(x). 
   \label{eq:architecture_d}
\end{equation}
Then, the proposed AdversarialNAS aims to learn a set of continuous parameters $\alpha$ and $\beta$ in a differentiable manner and obtain the final architecture of generator by simply preserving the most likely operations in the search space. Note that, we term the networks with all operations softly combined by the architecture parameters as Super-G and Super-D. The topology of the network would be changed by the learned high probability `None' operation, and the `Identity' operation would provide multi-scale fusion.

\subsection{Adversarial Architecture Search}
Before introducing the optimization strategy of the proposed AdversarialNAS, we first briefly revisit the optimization function in the classification literature. The searching process is formulated as a bilevel optimization problem:
\begin{equation}
   \begin{aligned}
      &\min\limits_{\alpha} \quad L_{val}(w^*(\alpha), \alpha)  \\
      s.t. \quad &w^*(\alpha)=\mathop{\arg\min}_{w} L_{train}(w,\alpha),
   \end{aligned}
      \label{eq:classification}
\end{equation}
where $L_{val}$ and $L_{train}$ denote the loss function on the validation and training set respectively. The goal of the search algorithm is to discover an optimal architecture $\alpha^*$ by calculating and minimizing the validation loss $L_{val}(w^*,\alpha)$, where $w^*$ is the optimal weights of the current architecture $\alpha$ and is obtained by calculating and minimizing the training loss $L_{train}(w,\alpha)$. Both the weight and architecture are optimized by ascending its gradient descent.

However, in the task of unconditional image generation, there are no labels to supervise the searching procedure. AutoGAN~\cite{gong2019autogan} and AGAN~\cite{wang2019agan} apply IS to evaluate the architecture performance and optimize the architecture by RL strategy. Computing IS requires generating hundreds of images and adopts Inception model to infer the result offline after a prolonged training trajectory of each discrete architecture, which is extremely time consuming. Therefore, we propose to make the architectures of generator and discriminator compete with each other to improve both of their performance, i.e., utilizing discriminator to guide the generator search and vice versa. AdversarialNAS leverages an adversarial optimization strategy that is inspired by the formulation of original GANs~\cite{goodfellow2014generative} for optimizing the architecture in a differentiable way. Thus the optimization process is defined as a two-player min-max game with value function $V(\alpha,\beta)$ where the weight of each network must be current optimal. The formulation of the introduced algorithm is given in Eqn.(~\ref{eq:optimization}):
\begin{equation}
   \begin{aligned}
      &\min\limits_{\alpha}\max\limits_{\beta}V(\alpha,\beta)=\mathbb{E}_{x\sim p_{data}(x)}[\log D(x\mid \beta,W_D^{*}(\beta))] \\
      &+\mathbb{E}_{z\sim p_z(z)}[\log (1-D(G(z\mid \alpha, W_G^*(\alpha))\mid \beta,W_D^*(\beta)))] \\ 
      \\
      &s.t. \\
      &W_D^*(\beta)=\mathop{\arg\max}_{W_D(\beta)}\mathbb{E}_{x\sim p_{data}(x)}[\log D(x\mid \beta,W_D(\beta))] \\
      &+\mathbb{E}_{z\sim p_z(z)}[\log(1-D(G_{D_\beta}^*(z)\mid \beta,W_D(\beta)))] \\
      &W_G^*(\alpha)=\mathop{\arg\min}_{W_G(\alpha)}\mathbb{E}_{z\sim p_z(z)}[\log(1-D_{G_\alpha}^*(G(\alpha\mid W_G(\alpha)))],
   \end{aligned}
      \label{eq:optimization}
\end{equation}
where $p_{data}$ means true data distribution and $p_z$ is a prior distribution. In the up-level stage the $W_D^*(\beta)$ denotes the optimal weights of discriminator under the specific architecture $\beta$ and $W_G^*(\alpha)$ represents the optimal weights of generator under the architecture $\alpha$. In the low-level stage, the two optimal weights $\{W_G^*(\alpha), W_D^*(\beta)\}$ for any particular pair of architectures $\{\alpha,\beta\}$ can be obtained through another min-max game between $W_G$ and $W_D$:
\begin{equation}
   \begin{aligned}
      &\min\limits_{W_G(\alpha)}\max\limits_{W_D(\beta)}V(W_G(\alpha),W_D(\beta))= \\
      &\mathbb{E}_{x\sim p_{data}(x)}[\log D(x\mid \beta,W_D(\beta))] \\
      &+\mathbb{E}_{z\sim p_z(z)}[\log (1-D(G(z\mid \alpha, W_G(\alpha))\mid \beta,W_D(\beta)))].
   \end{aligned}
      \label{eq:st}
\end{equation}

However, this inner optimization (Eq.~\ref{eq:st}) is time-consuming. For NAS in the classification task~\cite{liu2018darts,dong2019one,chen2019progressive}, the inner optimization (Eq.~\ref{eq:classification}) is normally approximated by one step training as $\nabla_{\alpha} L_{val}(w^*(\alpha), \alpha)\approx \nabla_{\alpha} L_{val}(w-\xi \nabla_{w} L_{train}(w,\alpha),\alpha)$. Inspired by this technique, for a given pair of architectures $\{\alpha,\beta\}$, the corresponding optimal weights $\{W_G^*(\alpha), W_D^*(\beta)\}$ can be obtained by single step of adversarial training (Eq.~\ref{eq:st}) as vanilla GANs. 
\begin{algorithm}[h]
   \caption{Minibatch stochastic gradient descent training of Adversarial Neural Architecture Search.}
   \begin{algorithmic}[1]
   \FOR{number of training iterations}
      \FOR{$k$ step}
         \STATE Sample minibatch of $2m$ noise samples $\left \{z^{(1)},...,z^{(2m)}  \right \}$ from noise prior.
         \STATE Sample minibatch of $2m$ examples $\left \{x^{(1)},...,x^{(2m)}  \right \}$ from real data distribution.
         \STATE Update the architecture of discriminator by ascending its stochastic gradient: \newline
         $\nabla_{\beta}\frac{1}{m}\sum_{i=1}^{m}\left [ \log(x^i)+\log(1-D(G(z^i))) \right ]$
         \STATE Update the weights of discriminator by ascending its stochastic gradient: \newline
         $\nabla_{W_D}\frac{1}{m}\sum_{i=m+1}^{2m}\left [ \log(x^i)+\log(1-D(G(z^i))) \right ]$
      \ENDFOR
   \STATE Sample minibatch of $2m$ noise samples $\left \{z^{(1)},...,z^{(2m)}  \right \}$ from noise prior.
   \STATE Update the architecture of generator by descending its stochastic gradient: \newline
   $\nabla_{\alpha}\frac{1}{m}\sum_{i=1}^{m}\left [\log(1-D(G(z^i))) \right ]$
   \STATE Update the weights of generator by descending its stochastic gradient: \newline
   $\nabla_{W_G}\frac{1}{m}\sum_{i=m+1}^{2m}\left [\log(1-D(G(z^i))) \right ]$
   \ENDFOR
   \end{algorithmic}
   \label{algorithm:adsearch}
\end{algorithm}
\vspace{-2.5mm}

Moreover, the min-max game between two architectures can also be searched in an alternative way. Specifically, the currently optimal architecture of generator for the given discriminator can be achieved through single step of adversarial training, which has been proven by Goodfellow in~\cite{goodfellow2014generative}. The proposed AdversarialNAS algorithm is shown in Alg.~\ref{algorithm:adsearch}, and optimal architectures or weights in each iteration can be achieved by ascending or descending the corresponding stochastic gradient. Note that, the order of the updating strategy is architecture first in each training iteration, which guarantees the weights for updating the corresponding architecture to be currently optimal. For example, the discriminator used in ninth line of Alg.~\ref{algorithm:adsearch} is D$^*$ with optimal architecture and weights for the current generator.

The proposed AdversarialNAS method can be plug-and-play to the original training procedure of GANs to search the architecture more naturally, which is specifically tailored for GANs.

\vspace{-1mm}
\section{Experiments}

\subsection{Experimental Setup}
\noindent\textbf{Datasets.} Following~\cite{gong2019autogan,wang2019agan}, we adopt CIFAR-10~\cite{krizhevsky2009learning} and STL-10 to evaluate the effectiveness of our approach. The CIFAR-10 contains 60,000 natural images including 10 different classes in $32\times 32$ spatial resolution. Concretely, we use its training set that consists of 50,000 images without any data augmentation technique to search for the optimal architecture of the generator. We also apply this training set to train the discovered architecture. To further evaluate the transferability of the architecture, we also adopt totally 105,000 images in STL-10 dataset to directly train the searched architecture without any data augmentation for a fair comparison with previous works. \newline
\textbf{Implementation.} We use Adam optimizer~\cite{kingma2014adam} and hinge loss to train the shared weights of Super-GAN and provide the supervision signal for updating the architectures. Specifically, the hyper-parameters of optimizers for training the weights of both generator and discriminator are set to $\beta_1=0.0$, $\beta_2=0.9$ and learning rate is set to 0.0002. The hyper-parameters of optimizers for optimizing both architectures are set to $\beta_1=0.5$, $\beta_2=0.9$ and the learning rate is 0.0003 with the weight decay of 0.0001. When searching, the batch size is set to 100 for both generator and discriminator, and we search for about 2,500 iterations. When training the derived generator, we directly adopt the discriminator used in AutoGAN~\cite{gong2019autogan} for a fair comparison, which is similar to the one in SNGAN~\cite{miyato2018spectral}. The batch size is set to 40 for generator and 20 for discriminator, respectively. We train the network for about 500 epochs, and the hyper-parameters of the optimizer are the same as the ones in searching. Besides, the same as all other methods, we randomly generate 50,000 images for calculating the Inception Score and FID to evaluate the network performance.\newline
\textbf{Computational Costs.} The proposed AdversarialNAS takes about 12 hours to converge for searching for an optimal architecture on two NVIDIA RTX 2080Ti GPUs. It requires only 1 GPU day to achieve the final architecture in a large search space (about $10^{38}$), while AutoGAN~\cite{gong2019autogan} requires 2 GPU days in a quite small search space (about $10^{5}$) and AGAN~\cite{wang2019agan} needs even 1200 GPU days for searching in a comparable space. Note that we directly use the released code of AutoGAN to search on the same hardware 2080Ti GPU and the searching time of AGAN is from their original paper (running on NVIDIA Titan X GPU).

\subsection{Compared with State-of-the-Art Approaches}
In this section, we discuss the searched architecture and compare its performance with state-of-the-art approaches including hand-crafted and auto-discovered ones. To explore the transferability of the discovered architecture, we directly apply it to another dataset and retrain its weights for comparing with other methods. Besides, we further study the scalability of the searched architecture and prove its superiority to other methods.

\vspace{-1.5mm}
\subsubsection{Results on CIFAR-10}
At the end of the searching program, we directly sample the architecture from the search space by picking the operations with maximum weights $\alpha$. The optimal architecture searched on CIFAR-10 is shown in Tab.~\ref{table:architecture} and some valuable observations can be received from this table.
\vspace{-1.5mm}
\begin{itemize}
   \item The searched generator prefer `Bilinear' operation for upsampling features although it has no learnable parameters. Besides, the `Bilinear Interpolation' provides more accurate expanded features than simple `Nearest ' operation, which is discovered by the searching algorithm.
   \vspace{-1mm}
   \item Surprisingly, there is no dilated convolution in this architecture. It seems that, for low-resolution images ($32\times 32$), simply stacking normal convolutions may already satisfy and achieve the optimal Effective Receptive Field (ERF) of the generator.
   \vspace{-1mm}
   \item We can also observe that the deeper cell tends to be more shallow since more `None' operations are preferred. The shallow cell has more multi-scale feature fusion operation, which is represented by the discovered parallel `Identity' connection of convolution.
\end{itemize}
\vspace{-1.5mm}
The quantitative comparisons with previous state-of-the-art methods are given in Tab.~\ref{table:comparisons}. From the table, we can see that the proposed AdversarialNAS is the first gradient-based approach that can search in a large search space with affordable cost. The designed search space has $10^{38}$ different architectures of GANs, which is several orders of magnitude larger than the search space ($10^{5}$) of AutoGAN~\cite{gong2019autogan}. Moreover, the proposed method only takes about 1 GPU day for searching for an optimal architecture while the AGAN~\cite{wang2019agan} spends 1200 GPU days under a comparable search space. In the CIFAR-10 dataset, our discovered `AdversarialNAS-GAN' achieves new state-of-the-art FID score ($10.87$), which is quite encouraging. It also obtains an Inception Score ($8.74\pm 0.07$) that is highly competitive with state-of-the-art Progressive GAN~\cite{karras2017progressive} ($8.80\pm 0.05$) and superior to AutoGAN~\cite{gong2019autogan} ($8.55\pm 0.10$). It is worth noting that the Progressive GAN applies a well-designed progressive training strategy that is time-consuming, while we directly train the discovered generator as vanilla GANs. 

\begin{figure}[b]
   \vspace{-3.5mm}
   \begin{center}
   \includegraphics[width=1\linewidth]{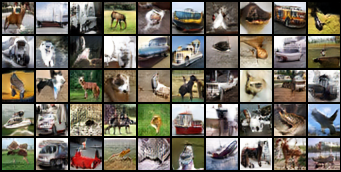}
\end{center}
   \vspace{-2mm} 
   \caption{The CIFAR-10 images generated by discovered generator in random without cherry-picking.}
   \label{fig:cifar}
   \vspace{-3mm}
\end{figure}

Besides, we randomly generate 50 images without cherry-picking, which are given in the Fig.~\ref{fig:cifar}. These qualitative results demonstrate that our searched generator can create diverse images that contain realistic appearance and natural texture without any clue of model collapse.

\begin{table}
   \begin{center}
   \begin{tabular}{|l|l|l|l|l|}\hline
   Up-Cell                       & Edge      & Operation        & Num  & Resolution             \\\hline
   \multirow{8}{*}{Cell-1} & $0\rightarrow 1$& Bilinear        & $1$  & $4\rightarrow 8$  \\
                           & $0\rightarrow 2$& Bilinear        & $1$  & $4\rightarrow 8$    \\ \cline{2-5}
                           & $1\rightarrow 3$& Identity        & $1$  & $8\rightarrow 8$       \\
                           & $1\rightarrow 4$& Conv $3\times 3$& $256$& $8\rightarrow 8$        \\ \cline{2-5}
                           & $2\rightarrow 3$& None            & $-$  & $-$                     \\
                           & $2\rightarrow 4$& Conv $3\times 3$& $256$& $8\rightarrow 8$        \\\cline{2-5}
                           & $3\rightarrow 4$& Identity        & $1$  & $8\rightarrow 8$        \\
                           & $3\rightarrow c_2$& Bilinear      & $1$  & $8\rightarrow 16$     \\
                           & $3\rightarrow c_3$& Nearest       & $1$  & $8\rightarrow 32$      \\\hline\hline
   \multirow{8}{*}{Cell-2} & $0\rightarrow 1$& Bilinear        & $1$  & $8\rightarrow 16$  \\
                           & $0\rightarrow 2$& Bilinear        & $1$  & $8\rightarrow 16$        \\\cline{2-5}
                           & $1\rightarrow 3$& None            & $-$  & $-$                     \\
                           & $1\rightarrow 4$& Conv $3\times 3$& $256$& $16\rightarrow 16$       \\ \cline{2-5}
                           & $2\rightarrow 3$& Identity        & $1$  & $16\rightarrow 16$     \\
                           & $2\rightarrow 4$& Conv $3\times 3$& $256$& $16\rightarrow 16$       \\\cline{2-5}
                           & $3\rightarrow 4$& Conv $3\times 3$& $256$& $16\rightarrow 16$      \\
                           & $3\rightarrow c_3$& Nearest       & $1$  & $16\rightarrow 32$     \\\hline\hline
   \multirow{8}{*}{Cell-3} & $0\rightarrow 1$& Nearest         & $1$  & $16\rightarrow 32$  \\
                           & $0\rightarrow 2$& Bilinear        & $1$  & $16\rightarrow 32$   \\\cline{2-5}
                           & $1\rightarrow 3$& None            & $-$  & $-$                   \\
                           & $1\rightarrow 4$& Conv $3\times 3$& $256$& $32\rightarrow 32$      \\ \cline{2-5}
                           & $2\rightarrow 3$& Conv $3\times 3$& $256$& $32\rightarrow 32$      \\
                           & $2\rightarrow 4$& None            & $-$  & $-$                      \\\cline{2-5}
                           & $3\rightarrow 4$& Conv $3\times 3$& $256$& $32\rightarrow 32$     \\\hline                      
   \end{tabular}
\end{center}
   \caption{The searched optimal architecture of generator by the proposed AdversarialNAS on CIFAR-10 with no category labels used. The `Num' indicates the number of operations.}
   \label{table:architecture}
   \vspace{-3mm}
\end{table}

\begin{table*}[t!]
   \begin{center}
   \begin{tabular}{| l | c | c | c | c | l | l | l | l |} \hline
   Method & \makecell{Search\\Method} & \makecell{Search\\Space} &  \makecell{Search\\Cost}  & \makecell{Size\\(MB)} & \makecell{IS$\uparrow$ on\\ C-10} & \makecell{FID$\downarrow$ on \\ C-10} & \makecell{IS$\uparrow$ on \\ S-10} & \makecell{FID$\downarrow$ on \\ S-10}  \\
   \hline
   DCGANs~\cite{radford2015unsupervised}            &       &    &    &     & $6.64\pm 0.14$     & $-$    & $-$            & $-$        \\
   Improved GAN~\cite{salimans2016improved}         &       &    &    &     & $6.86\pm 0.06$     & $-$    & $-$            & $-$        \\
   LRGAN~\cite{yang2017lr}                          &       &    &    &     & $7.17\pm 0.17$     & $-$    & $-$            & $-$        \\
   DFM~\cite{WardeFarley2017ImprovingGA}            &       &    &    &     & $7.72\pm 0.13$     & $-$    & $8.51\pm 0.13$ & $-$        \\
   ProbGAN~\cite{He2019ProbGANTP}                   &       &    &    &     & $7.75$             & $24.6$ & $8.87\pm 0.09$ & $46.74$   \\
   WGAN-GP, ResNet~\cite{Gulrajani2017ImprovedTO}   & Manual& $-$& $-$&  $-$& $7.86\pm 0.07$     & $-$    & $-$            & $-$        \\
   Splitting GAN~\cite{Grinblat2017ClassSplittingGA}&       &    &    &     & $7.90\pm 0.09$     & $-$    & $-$            & $-$        \\ 
   MGAN~\cite{Hoang2018MGANTG}                      &       &    &    &     & $8.33\pm 0.10$     & $26.7$ & $-$            & $-$        \\
   Dist-GAN~\cite{Tran2018DistGANAI}                &       &    &    &     & $            $     & $17.61$& $-$            & $36.19$  \\
   Progressive GAN~\cite{karras2017progressive}     &       &    &    &     & $\bf{8.80\pm 0.05}$& $-$    & $-$            & $-$        \\
   Improving MMD-GAN~\cite{Wang2018ImprovingMT}     &       &    &    &     & $8.29        $     & $16.21$& $9.23\pm 0.08$ & $37.64$ \\
   SN-GAN~\cite{miyato2018spectral}                 &       &    &    &$4.3$& $8.22\pm 0.05$     & $21.7$ & $9.16\pm 0.12$ & $40.1$   \\
   \hline\hline
   AGAN~\cite{wang2019agan}                         & RL    & $-$   & $1200$& $20.1$& $8.29\pm 0.09$& $30.5$ & $9.23\pm 0.08$& $52.7$   \\
   Random Search~\cite{li2019random}$\dagger$       & Random& $10^5$& $2$   & $-$   & $8.09$        & $17.34$& $-$           & $-$    \\
   AutoGAN~\cite{gong2019autogan}                   & RL    & $10^5$& $2$   & $4.4$ & $8.55\pm 0.10$& $12.42$& $9.16\pm 0.12$& $31.01$  \\
   \hline
   Random Search~\cite{li2019random}$\dagger\dagger$ & Random   & $10^{38}$& $1$& $12.5$& $6.74\pm 0.07$& $38.32$     & $7.66\pm 0.08$     & $53.45$    \\
   AdversarialNAS-GAN                                & Gradient & $10^{38}$& $1$& $8.8$ & $8.74\pm 0.07$& $\bf{10.87}$& $\bf{9.63\pm 0.19}$& $\bf{26.98}$ \\
   \hline
   \end{tabular}
\end{center}
   \caption{The quantitative comparisons with state-of-the-art approaches. $\dagger$ indicates the results are achieved in the search space of AutoGAN and $\dagger\dagger$ denotes the results in our search space.}
   \label{table:comparisons}
   \vspace{-2mm}
   \end{table*}

\subsubsection{Transferability of the Architectures}
Following the setting of AutoGAN~\cite{gong2019autogan} and AGAN~\cite{wang2019agan}, we directly apply the generator searched on CIFAR-10 to STL-10 dataset for evaluating the transferability of the architecture. Specifically, we adopt totally 105,000 images with no labels used to train this network. The number of training epochs is the same as the one on CIFAR-10 and we also randomly generate 50,000 images for calculating the Inception Score and FID. We alter the resolution of input noise to $6\times 6$ for generating the image with the size of $48\times 48$, as the AutoGAN and AGAN do. 

\begin{figure}
   \begin{center}
   \includegraphics[width=1\linewidth]{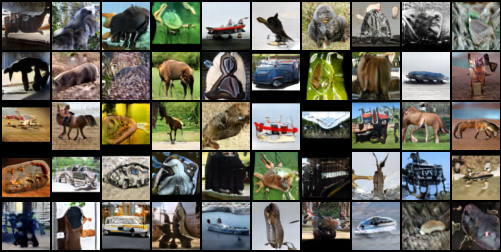}
\end{center} 
   \vspace{-2mm}
   \caption{The STL-10 images randomly generated without cherry-picking by the generator discovered on CIFAR-10.}
   \label{fig:stl}
   \vspace{-3mm}
\end{figure}

The quantitative results are shown in Tab.~\ref{table:comparisons}. We can observe that our network suffers no overfitting on the CIFAR-10 dataset and has a superior ability of generalization. Specifically, it achieves the state-of-the-art Inception Score ($9.63$) and FID ($26.98$) on STL-10, which are far better than all hand-crafted and auto-discovered methods. The qualitative results are also given in Fig.~\ref{fig:stl} to prove its ability to generate diverse and realistic images.

\begin{table*}
   \begin{center}
   \begin{tabular}{lcccccc}
   \toprule
   \multirow{2}{*}{Methods} & \multicolumn{2}{c}{Discriminator} & \multicolumn{2}{c}{CIFAR-10}      & \multicolumn{2}{c}{STL-10}       \\\cmidrule{4-7}
                            & Architecture  & Type              & IS$\uparrow$   & FID$\downarrow$   & IS$\uparrow$    & FID$\downarrow$     \\\midrule
   Random Search           & Fixed        & AutoGAN-D     & $6.74\pm 0.13$     & $38.32$       & $7.66\pm 0.11$      & $53.45$           \\\midrule
   SingalNAS               & Fixed         & SNGAN-D      & $7.72\pm 0.03$      & $27.79$      & $6.56\pm 0.12$       & $84.19$          \\
   SingalNAS               & Fixed        & AutoGAN-D       & $7.86\pm 0.08$    & $24.04$       & $8.52\pm 0.05$       & $38.85$           \\
   SingalNAS               & Fixed           & Super-D      & $7.77\pm 0.05$    & $23.01$      & $8.62\pm 0.03$       & $41.57$               \\\midrule
   AdversarialNAS           & Dynamic       & Searched-D     & $\bf{8.74\pm 0.07}$  & $\bf{10.87}$    & $\bf{9.63\pm 0.19}$ & $\bf{26.98}$           \\\bottomrule
   \end{tabular}
\end{center}
   \caption{We search the generative model on CIFAR-10 with different methods and retain the weight of these searched architectures to evaluate their performance on both CIFAR-10 and STL-10.}
   \label{table:ablation}
\end{table*}

\subsubsection{Scalability of the Architectures}
In this section, we further explore the scalability of the discovered architecture on the CIFAR-10 dataset.

\begin{figure}[b]
   \vspace{-2.5mm}
   \begin{center}
   \includegraphics[width=1\linewidth]{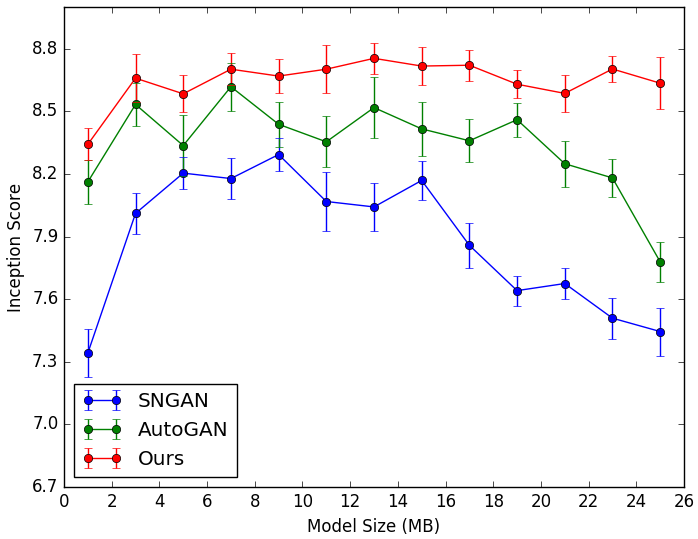}
\end{center} 
   \caption{Inception Score curves of different methods.}
   \label{fig:IS}
\end{figure}

\begin{figure}
   \begin{center}
   \includegraphics[width=1\linewidth]{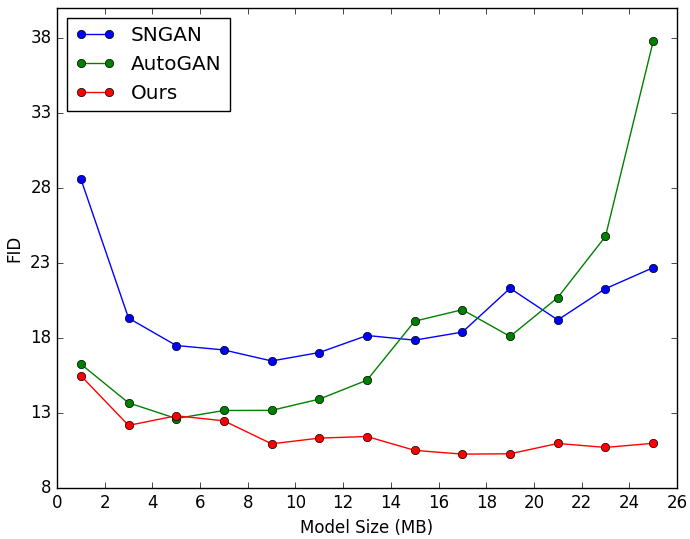}
\end{center} 
   \caption{FID curves of different methods.}
   \label{fig:FID}
   \vspace{-4mm}
\end{figure}

We compare our searched generator with two representative works, manual-designed SNGAN~\cite{miyato2018spectral} and auto-discovered AutoGAN~\cite{gong2019autogan}. 
We scale the parameter size of these generators from 1 MB to 25 MB through channel dimension, which is a large scope. Note that, for a fair comparison, we use the same discriminator with a fixed size in all experiments to observe the impact of generator capacity changes. The qualitative comparisons are illustrated in Fig.~\ref{fig:IS} and Fig.~\ref{fig:FID}. The x-axis in both figures denotes the parameter size (MB) of the specific generator. The y-axis is IS in Fig.~\ref{fig:IS} and is FID in Fig.~\ref{fig:FID}. These experiments demonstrate that our searched architecture is more stable and almost unaffected when scaling the model size. When the size is extremely compressed to only $1$ MB, the SNGAN and AutoGAN all suffer from the disaster of performance degradation, while the performance of `AdversarialNAS-GAN' is almost unchanged. Notably, the performance will continually drop when expanding the generator size because the enlarged generator will not be balanced with the fixed-size discriminator any more. However, both Fig.~\ref{fig:IS} and Fig.~\ref{fig:FID} demonstrate that our discovered architecture will not suffer from performance drop, which means it is more robust and has superior inclusiveness for discriminators.


\subsection{Ablation Studies}
To further evaluate the effectiveness of the proposed AdversarialNAS, we conduct a series of ablation studies.

First, we conduct a random search strategy~\cite{li2019random} to search for the generator where we adopt the fixed-architecture discriminator of AutoGAN for a fair comparison. The performance of the searched generative model is shown in Tab.~\ref{table:ablation}. Second, we propose `SingalNAS' to search the optimal generator with different types of fixed architecture of discriminator, while the weights of discriminator can still be trained. Accordingly, the supervision signal for updating the generator architecture comes from the fixed architecture of discriminator, and the discriminator architecture does not dynamically change according to generator during searching. We adopt the discriminator architecture of SNGAN and AutoGAN, respectively. In addition, to verify the influences of our search space, we also conduct `SingalNAS' with the fixed Super-D. Third, we use the proposed `AdversarialNAS' to search the generator and discriminator simultaneously. Note that, the time consuming of both searching and training in all experiments is constrained to be consistent. 

The effectiveness of our adversarial searching strategy can be observed from the comparisons in Tab.~\ref{table:ablation}.




\section{Conclusion}
In this work, we propose a large search space for GANs and a novel AdversarialNAS method to search for a superior generative model automatically. The proposed searching algorithm can directly be inserted to the original procedure of GAN training and search the architecture of generator in a differentiable manner through an adversarial mechanism, which extremely reduces the search cost. The discovered network achieves state-of-the-art performance on both CIFAR-10 and STL-10 datasets, and it also has advanced transferability and scalability.

Furthermore, we believe the idea behind our AdversarialNAS is not only specific to NAS-GAN and may benefit other potential field where there are multiple network architectures requiring mutual influence, such as network architecture distillation, pruning and mutual learning.

{\small
\bibliographystyle{ieee_fullname}
\bibliography{egbib}
}

\end{document}